\pdfoutput=1
\documentclass[letterpaper, 10 pt, conference]{ieeeconf}  

\IEEEoverridecommandlockouts                              
\usepackage{amssymb}
\usepackage{graphicx}
\usepackage{cite}
\usepackage{color} 
\usepackage{fancybox}
\usepackage{amssymb}
\usepackage{enumerate}
\usepackage{graphicx}
\usepackage{amsmath}
\usepackage{algpseudocode}
\usepackage{algorithm}
\usepackage{graphicx,multirow}

\newcommand{\Real}{\mathbb{R}}
\newcommand{\Rn}{\mathbb{R}^n}
\newcommand{\hamid}[1]{\textcolor{black}{#1}}
\newcommand{\reza}[1]{\textcolor{black}{#1}}

\usepackage[labelfont=bf,labelsep=period,justification=raggedright]{caption}

\title{\LARGE \bf
NUROA: A Numerical Roadmap Algorithm
}

\author{Reza Iraji and Hamidreza Chitsaz
\thanks{R. Iraji and H. Chitsaz are with the Department of Computer Science,
        \hamid{Colorado State University, 1100 Central Avenue Mall, Fort Collins, CO
        {\tt\small rezairaji@gmail.com, chitsaz@chitsazlab.org}}}}

\begin{document}

\maketitle
\thispagestyle{empty}
\pagestyle{empty}

\begin{abstract}
Motion planning has been studied for nearly four decades now.
Complete, combinatorial motion planning approaches are theoretically well-rooted with completeness guarantees but they are hard to implement. Sampling-based and heuristic methods are easy to implement and quite simple to customize but they lack completeness guarantees. Can the best of both worlds be ever achieved, particularly for mission critical applications such as robotic surgery, space explorations, and handling hazardous material? In this paper, we answer affirmatively to that question. We present a new methodology, NUROA, to numerically approximate the Canny's roadmap, which is a network of one-dimensional algebraic curves. Our algorithm encloses the roadmap with a chain of tiny boxes each of which contains a piece of the roadmap and whose connectivity captures the roadmap connectivity. It starts by enclosing the entire space with a box. In each iteration, remaining boxes are shrunk on all sides and then split into 
smaller sized boxes. Those boxes that are empty are detected in the shrink phase and removed. The algorithm terminates when all remaining boxes are smaller than a resolution that can be either given as input or automatically computed using root separation lower bounds. Shrink operation is cast as a polynomial optimization with semialgebraic constraints, which is in turn transformed into a (series of) semidefinite programs (SDP) using the Lasserre's approach. NUROA's success is due to fast SDP solvers. NUROA correctly captured the connectivity of multiple curves/skeletons whereas competitors such as IBEX and Realpaver failed in some cases. Since boxes are independent from one another, NUROA can be parallelized particularly on GPUs. NUROA is available as an open source package at {\tt http://nuroa.sourceforge.net/}.
\end{abstract}

\section{Introduction}
With the advent of automated manufacturing and robotics, the field of motion planning was introduced to the scientific society by the pioneering works of Lozano-Perez and Reif \cite{LozWes79,Rei79}. A robot usually works in a 2D or 3D environment, called work space, containing obstacles. Lozano-Perez suggested that a layer of abstraction can be added by associating any motion of the robot with a path in the set of feasible distinct robot configurations, also known as the configuration space $C$. That association induces a natural correspondence between work space obstacles $O$ and obstacle regions in the configuration space $C_{obs}$. Often, the collision-free subset of the configuration space $C_{free} = C \backslash C_{obs}$ can be explained by a set of polynomial inequalities that are computed from the description of $O$ and robot. The input of a motion planning problem is that set of polynomial inequalities and an initial and a goal point in $C_{free}$.

Early in the field, the motion planning problem was proved to be PSPACE-hard and consequently NP-hard \cite{Rei79}. In the first attempts to solve the problem, researchers aimed at complete, combinatorial algorithms. This led to some outstanding works such as the use of Collins cylindrical algebraic decomposition by Schwartz and Sharir \cite{Col75,Collins76,SchSha83b,SchSha83c,SchSha83} and the Canny's roadmap algorithm \cite{Can88}.

Schwartz and Sharir gave the first complete motion planning algorithm for a rigid body in two and three dimensions \cite{SchSha83b,SchSha83c,SchSha83}. Their algorithm is based on algebraic geometry methods, specifically cylindrical algebraic decomposition \cite{Col75,Collins76}. The running time of Schwartz-Sharir algorithm is doubly-exponential in the dimension of the configuration space. Canny introduced a singly exponential time complexity algorithm based on the Morse theory and resultants in commutative algebra, which is near optimal provided $P \neq NP$. Recently, the Canny's algorithm was improved by Basu \emph{et al.} \cite{BasPolRoy00,BasPolRoy06,BasRoySafSch12}. Safey El Din and Schost have embarked on a journey to achieve the optimal roadmap algorithm by a (nearly) balanced division of dimension at each recursive iteration of the algorithm \cite{Din13}.

\begin{figure}[t!]
\begin{center}
\begin{tabular}{c}
\includegraphics[width=0.98\columnwidth]{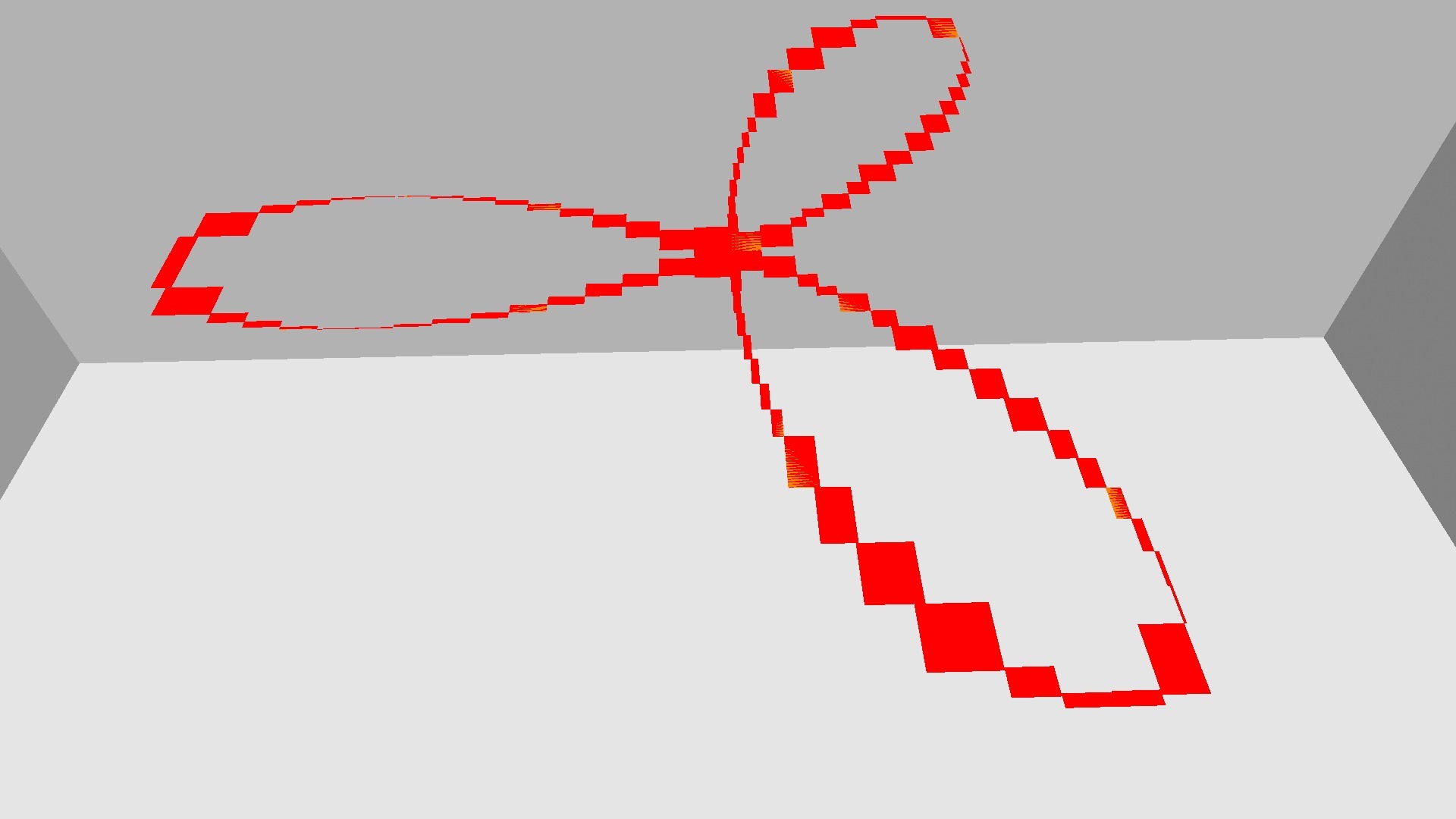} \\
Three-leaf clover planar curve\\
\end{tabular}
\end{center}
\caption{A boxified three-leaf clover planar curve in $\Real^3$.}\label{fig:clover}
\end{figure}

Although the Canny's algorithm was the theoretical bottom line, it was of little use in practice since implementing it involved an unmanageable level of sophistication. To the best of our knowledge, there is still no publicly available implementation of the roadmap algorithm to date. That fact inspired another trend in the field in the 1990s to address practical motion planning problems. Discretization and grid search were among the first attempts along those lines \cite{CheHwa92,Kon91}. Numerous easy to implement heuristic approaches such as artificial potential fields \cite{Kha86,LoiDimKyr04,iraji07,iraji09} and sampling-based motion planning algorithms that claim probabilistic completeness \cite{Kav94,Lavalle00} have appeared. 

For some non-critical applications, sampling-based motion planning algorithms have proven to be applicable in practice. \hamid{Even though for some critical applications such as surgery and nuclear material handling, heuristic motion planners have been deployed \cite{Alterovitz08,Kineo}, one cannot generally assume that heuristic motion planning for many other mission critical applications  such as space explorations and nuclear facility repair can be used.} We conjecture that if robots are to ubiquitously enter our day-to-day lives, they have to be equipped with advanced, theoretically well-rooted motion planners with some form of completeness guarantee.
 
At first glance, completeness and practicality may appear unattainable at the same time. Complete general motion planning algorithms, namely cylindrical algebraic decomposition and the roadmap algorithm, are based on real algebraic geometry computations which is extremely difficult to implement. Although the problem was theoretically solved by the Canny's innovative algorithm, it remains open from a practical perspective. The reason is that solving the Canny's polynomial system of equations is intractable symbolically. More precisely, simplifying that system of equations into one polynomial equation, called the resultant, is very hard. 

\reza{An attempt has been made to alleviate this hardness by employment of numerical computation of critical points of the roadmap \cite{HirMouPap00}. In comparison with our method, which computes the whole roadmap numerically, that work proposed an algorithm to bypass symbolic computation of the determinants of resultant matrices by numerical determination of just the 
critical, turning, and self-crossing points. Although that idea slightly improved practicality of the Canny's algorithm, the whole roadmap algorithm still remained impractical.}

Furthermore, even though sampling-based motion planning algorithms became popular due to solving the problem for some non-critical applications, they never met the reliability expectations for mission-critical applications. The convergence rate of sampling-based algorithms is low in the case of narrow passages, which occur frequently in real world. Sampling-based algorithms often waste computation on the wrong part of a problem, namely narrow passages that do not even cause NP-hardness \cite{SchShaHop86}. Also, local approaches such as artificial potential field suffer from getting trapped in local minima. Dealing with this problem, either by designing a navigation function which guarantees no local minima \cite{Choset04} or by 
heuristic approaches \cite{iraji07}, makes the problem so complicated that sometimes cannot be solved at all or makes it unreliable and consequently unsuitable for mission critical applications.

Despite decades of research in motion planning, there is still a gap in the field. To fill this gap, we introduce a new methodology, inspired by the roadmap algorithm, in this paper to simultaneously maintain both resolution-completeness and practicality for a large class of problems.


\section{Problem}\label{sec:problem}
We consider the problem of planning motion for robotic systems. Each robot is composed of arbitrary open/closed kinematic chains of bodies (generally semialgebraic objects). Two objects are in collision if their surfaces are closer than $\Delta$ in the Euclidean space. Mathematically, the space of collision-free configurations of the entire system of robots ($C_{free}$) can be represented by a semialgebraic set 
\begin{equation}
S := \{x \in \Rn\ |\ f_1(x) \geq 0, f_2(x) \geq 0, \ldots, f_m(x) \geq 0\},
\end{equation}
in which $f_i \in \Real[X]$ are polynomials (see Chapter 2 of \cite{Can88} for a detailed derivation of $f$ polynomials). Note that $S$ is parametrized by $\Delta$, but for the sake of brevity we do not explicitly denote it in this paper. We denote $X_1, X_2, \ldots, X_n$ variables by $X$. We are given the $f_i$ polynomials and the initial configuration $x_I$ and the goal configuration $x_G$ in $S$. The motion planning problem asks for a path in $S$ from $x_I$ to $x_G$ if there is such a path and non-existence report otherwise. The Canny's roadmap algorithm computes a one-dimensional semialgebraic subset of $S$, called the roadmap, and pieces of semialgebraic curves connecting $x_I$ and $x_G$ to the roadmap. A graph connectivity query from $x_I$ to $x_G$ yields the result. Since the roadmap is guaranteed to cross every Morse slice of the configuration space, it essentially captures the topology of the space. 


Our idea is to approximate edges of the roadmap, which are algebraic curves implicitly defined as the zero set of a system of polynomials, by chains of adjacent bounding boxes in $\Rn$, each of which contains a slice of the roadmap. Faces of such bounding boxes are computed by iterative polynomial optimization on semialgebraic sets \cite{Lasserre01,Schweighofer05} which are performed in turn by a series of semidefinite programs (SDP) \cite{VanBoy94}. Our approach is in the spirit of Porta's \emph{et al.} \cite{Porta07,PorRosTho09}, albeit using a more sophisticated SDP-based optimization on the roadmap. \hamid{Note that IBEX and Realpaver are two interval analysis and constraint satisfaction algorithms that can approximate algebraic curves by a collection of bounding boxes \cite{Araya14,Granvilliers06}. However, we will demonstrate in the results section that they compromise accuracy, as they deal with general objective and constraint functions not just polynomials.}

The roadmap algorithm consists of computing the roadmap for real algebraic varieties 
\begin{equation}\label{equ:g}
g(x) = f_{i_1}(x)^2 + f_{i_2}(x)^2 + \cdots + f_{i_k}(x)^2 = 0, 
\end{equation}
where $1 \leq k \leq \ell \leq m$ and $\ell$ is the Basu-Pollack-Roy complexity of $S$ \cite{BasPolRoy00}, and $\{i_1, i_2, \ldots, i_k\}$ ranges over all possible $k$-element subsets of $\{1, 2, \ldots, m\}$. These individual roadmaps are then connected along their intersections with other varieties to form the global roadmap of $S$. In reality, many of these varieties are empty in which case their roadmap is just empty. Our numerical algorithm detects such cases in the first step, which will cause the expected number of considered varieties in our algorithm much lower than the worst case exponential $m^\ell$. Moreover, modern collision detection techniques \cite{PanChiMan12} provide valuable information that can help detect such empty varieties beforehand. In the following, the approach is presented in more detail.

\section{Approach}\label{sec:approach}
There are two types of objects that are approximated numerically in our work: curves and points. Curves, which are edges of the roadmap mainly captured in the skeleton (see below), are represented by a chain of adjacent bounding boxes \cite{PorRosTho09}. A point is represented by one bounding box. Both curves and points arise as the zero sets of systems of polynomials that are computed in the roadmap algorithm. We first explain below how those systems of polynomials are computed, which is taken from the roadmap algorithm. We then explain how we approximate a one or zero dimensional zero set of a system of polynomials. Note that our approximation scheme preserves resolution-completeness of the roadmap algorithm. 

\subsection{Skeleton}
Within the core of the roadmap algorithm lies computation of the skeleton, which is the preimage of the silhouette of the projection of the input variety (\ref{equ:g}) onto the first two coordinates. More precisely, the skeleton is the first approximation of 
\begin{equation}\label{equ:skeleton}
\Sigma(\epsilon) := \{x \in \Rn\ |\ g(x) = \epsilon, \frac{\partial g}{\partial x_3} = 0, \ldots, \frac{\partial g}{\partial x_n} = 0 \}, 
\end{equation}
as $\epsilon \to 0$. The roadmap algorithm treats $\epsilon$ as a variable and employs elimination theory to compute a resultant polynomial $h \in \Real[X_1, X_2, \epsilon]$ such that 
\begin{equation}
\Sigma(\epsilon) = \{x \in \Rn\ |\ g(x) = \epsilon, h(x_1, x_2, \epsilon) = 0 \}. 
\end{equation}

The zero set of the coefficients of the lowest degree $\epsilon$ in $h(x_1, x_2, \epsilon)$ together with $g(x) = 0$ define $\Sigma$. For more general cases, advanced multi-infinitesimal-based algebraic methods have been given to compute the roadmap skeleton \cite{BasPolRoy00}. Our method is much easier to implement since we will use numerical calculations instead of computer algebra. 

\subsection{Points} 
A point $A \in S$ is called $X_1$-critical if $\frac{\partial g}{\partial x_2}|_{A} = 0$. In the roadmap algorithm \cite{Can88}, recursive calls to the skeleton algorithm are performed on the slices of $S$ at $X_1$-critical points. In other parts of the roadmap algorithm, intersections of $\Sigma$ with other varieties, which are gluing vertices of the roadmap, are computed. 

For all those points, our algorithm computes a bounding box, instead of an exact algebraic point (zero set of a resultant polynomial), using Newton method. For instance, bounding boxes are computed by intersecting the chain of bounding boxes in $\Sigma$ with the variety in the other leg of intersection.

\subsection{Lazy Recursions} 
Those recursive calls to the skeleton algorithm are slightly more complicated in our case. Let $A = (a_1, a_2, \ldots, a_n) \in \Rn$ be an $X_1$-critical point. The Canny's algorithm calls the skeleton algorithm on $S\ \cap\ \{x\in \Rn\ |\ x_1 = a_1\}$ slice of $S$. Our algorithm does not compute $A$ precisely, but it rather approximates $A$ numerically by a bounding box $[a^l_1, a^u_1]\times\cdots\times[a^l_n, a^u_n] \ni A$. At what slice should the roadmap algorithm be recursively called? 

The skeleton algorithm is called twice: once for the lower slice $S \cap \{x\in \Rn\ |\ x_1 = a^l_1\}$, and once for the upper slice $S \cap \{x\in \Rn\ |\ x_1 = a^u_1\}$. Both skeletons are then added to the roadmap. The Morse theorem shows that the topology of $X_1$-slices do not change between consecutive $X_1$-critical values \cite{Can88}. Hence, our algorithm guarantees resolution-completeness provided that exactly one $X_1$-critical value is contained within $[a^l_1, a^u_1]$ interval. \hamid{To achieve completeness, that resolution can be computed from root separation lower bounds that do not require symbolic computations \cite{Rump79,Mignotte83,Collins01}.} 

After completion of each recursive call to the skeleton algorithm, our algorithm searches the partially built roadmap to see if it finds a path from $x_I$ to $x_G$ in the roadmap. Often, a path may appear in partially built roadmaps, in which case our algorithm will not continue recursions further on $X_1$-critical intervals and will save computation time.

\subsection{Bounding Boxes} 
Given a skeleton piece (point) of the roadmap, which is a one (zero) dimensional real variety in $S$,
\begin{equation}\label{equ:variety}
Z := S\ \cap\ \{x \in \Rn\ |\ h_1(x) = 0, h_2(x) = 0, \ldots, h_k(x) = 0\},
\end{equation}
our algorithm computes a set of sufficiently small boxes $B$ that contain $Z$, i.e. $Z \subset \bigcup_{b\in B} b$. \hamid{For instance in the case of skeleton piece, $h_1(x) = g(x) - \epsilon$, $h_2(x) = \partial g/\partial x_3$, $h_3 = \partial g/\partial x_4$, etc.} The algorithm starts with an initial box set
\begin{equation}
B = \Bigl\{ [l_1, u_1]\times[l_2, u_2]\times\cdots\times[l_n, u_n] \Bigr\},
\end{equation}
containing the entire configuration space $S \subset [l_1, u_1]\times[l_2, u_2]\times\cdots\times[l_n, u_n]$. Non-compact configuration spaces can be compactified; hence, we assume $S$ is compact, in which case there is such an initial bounding box. Our algorithm iterates over two operations, \emph{shrinking} and \emph{splitting}, on bounding boxes in $B$. Shrinking eliminates portions of a box that do not contain any piece of the variety, and box splitting refines the resolution. This iterative process continues until all boxes are either empty or sufficiently small. Our algorithm is inherently multi-resolution, which means the termination criteria can be evaluated box by box, locally based on neighboring boxes, root separations \cite{Rump79,Mignotte83,Collins01}, and also based on criticality of application.

\subsubsection{Box Shrinking}\label{sec:shrinking} 
Given a box $b = [l_1, u_1]\times[l_2, u_2]\times\cdots\times[l_n, u_n]$, this module of the algorithm squeezes $b$ to obtain the smallest box $b'= [l'_1, u'_1]\times[l'_2, u'_2]\times\cdots\times[l'_n, u'_n] \subseteq b$ that contains $Z\cap b$. Our algorithm iteratively shrinks the interval of each dimension until no more shrinking is possible. Here, we present our algorithm to shrink $[l_i, u_i]$ to obtain $[l''_i, u''_i]$. Note that we cannot necessarily obtain $[l'_i, u'_i]$ in one step, and the algorithm iterates potentially multiple times over shrinking every dimension. However, the algorithm is able to discover empty boxes in one iteration.

Shrinking $[l_i, u_i]$ to obtain $[l''_i, u''_i]$ is cast as the following optimization problems
\begin{equation}\label{equ:opt}
\begin{aligned}
l''_i =\ & \underset{x}{\text{minimize}}
& & x_i \\
u''_i =\ & \underset{x}{\text{maximize}}
& & x_i \\
& \text{subject to} & & l_j \leq x_j \leq u_j, \; 1 \leq j \leq n, \\
&&& h_j(x) = 0, \; 1 \leq j \leq k, \\
&&& f_j(x) \geq 0, \; 1 \leq j \leq m, \\
&&& \left\lceil \sum_{j=1}^{n} l_j^2 + u_j^2 \right\rceil - \sum_{j=1}^{n} x_j^2 \geq 0,
\end{aligned}
\end{equation}
where the constraints correspond to the current bounding box, the input semialgebraic set (\ref{equ:variety}), and satisfaction of a technical assumption. We propose to solve these optimization problems using the Lasserre's approach \cite{Lasserre01,Schweighofer05} which requires satisfaction of a general assumption described below in (\ref{equ:assumption}) \cite{Schweighofer05}. That is why we added the last constraint above. Obviously, the last constraint does not affect the result. For the sake of presentation, let
\begin{align} \label{equ:e7}
c & = 2n+2k+m+1, \\ \label{equ:e8}
Z_b & = Z \cap b, \\
\label{equ:e0}
e_0(x) & = 1,\\ \label{equ:e1}
e_j(x) & = x_j-l_j, \; 1 \leq j \leq n, \\ \label{equ:e2}
e_{n+j}(x) & = u_j-x_j, \; 1 \leq j \leq n, \\ \label{equ:e3}
e_{2n+j}(x) & = h_j(x), \; 1 \leq j \leq k, \\ \label{equ:e4}
e_{2n+k+j}(x) & = -h_j(x), \; 1 \leq j \leq k, \\ \label{equ:e5}
e_{2n+2k+j}(x) & = f_j(x), \; 1 \leq j \leq m, \\ \label{equ:e6}
e_{2n+2k+m+1}(x) & = \left\lceil \sum_{j=1}^{n} l_j^2 + u_j^2 \right\rceil - \sum_{j=1}^{n} x_j^2.  
\end{align}
Using this notation, (\ref{equ:opt}) becomes
\begin{equation}\label{equ:opt-compact}
\begin{aligned}
l''_i =\ & \underset{x}{\text{minimize}}
& & x_i \\
u''_i =\ & \underset{x}{\text{maximize}}
& & x_i \\
& \text{subject to} & & e_j(x) \geq 0, \; 1 \leq j \leq c.
\end{aligned}
\end{equation}

Denote the set of all squares $p^2$ of polynomials $p \in \Real[X]$ by $\Real[X]^2$, the set of all $p^2e_j$ by $\Real[X]^2e_j$, and the set of all finite sums of such elements by $\sum \Real[X]^2e_j$. The set
\begin{equation}
\begin{split}
M &:=\sum \Real[X]^2 + \sum \Real[X]^2e_1 + \cdots + \sum \Real[X]^2e_c\\
& = \left\{ \sum_{j=0}^{c} q_je_j\ \Big|\ q_j \in \sum \Real[X]^2 \right\},
\end{split}
\end{equation}
is the quadratic module generated by $e_1, \ldots, e_c$. Note that addition of $e_c(x) \geq 0$ to the constraints helps satisfy the Lasserre's general assumption \cite{Schweighofer05}:
\begin{equation}\label{equ:assumption}
\exists N \in \mathbb{N} : N - \sum_{i=1}^{n} X_i^2 \in M, 
\end{equation}
with $N = \left\lceil \sum_{j=1}^{n} l_j^2 + u_j^2 \right\rceil$. The Lasserre's method convexifies the problem in two different ways. The first one is to exchange the points of the underlying feasible semialgebraic set $Z_b$ by probability measures on $Z_b$. Every point $x \in Z_b$ can be identified with the Dirac measure $\delta_x$ at $x$. Therefore, (\ref{equ:opt-compact}) is equivalent to
\begin{equation}\label{equ:opt-prim}
\begin{aligned}
l''_i & = \inf\left\{\int x_id\mu\ \Big|\ \mu \in \mathcal{M}^1(Z_b)\right\},\\
u''_i & = \sup\left\{\int x_id\mu\ \Big|\ \mu \in \mathcal{M}^1(Z_b)\right\},
\end{aligned}
\end{equation}  
in which $\mathcal{M}^1$ denotes the set of probability measures. The second method of convexification is to cast the dual problems as
\begin{equation}\label{equ:opt-dual}
\begin{aligned}
l''_i & = \sup\left\{a \in \Real \ |\ x_i - a > 0 \mbox{ on } Z_b\right\},\\
u''_i & = \inf\left\{a \in \Real \ |\ x_i - a < 0 \mbox{ on }
Z_b \right\}.
\end{aligned}
\end{equation}  

For the sake of brevity, we continue presenting our approach only for the lower bound $l''_i$. For the upper bound, we will use the obvious analogue. Using Putinar's \emph{Positivstellensatz} theorems \cite{Putinar93}, (\ref{equ:opt-prim}) becomes
\begin{equation}\label{equ:opt-prim-put}
\begin{split}
l''_i = \inf\{L(X_i)\ |\ L : \Real[X] \to \Real \mbox{ is linear, } L(1) = 1,\\
L(M) \subseteq [0, \infty) \},
\end{split}
\end{equation}
and (\ref{equ:opt-dual}) becomes
\begin{equation}\label{equ:opt-dual-put}
l''_i = \sup\left\{a \in \Real \ |\ X_i - a \in M \right\}.
\end{equation}

The idea is to relax (\ref{equ:opt-prim-put}) and (\ref{equ:opt-dual-put}) by approximations $M_d \subseteq \Real[X]_d$ of $M \subseteq \Real[X]$, in which $\Real[X]_d$ denotes the vector space of polynomials $p \in \Real[X]$ of degree at most $d$. More precisely,
\begin{equation}
\begin{split}
M_d &:=\sum \Real[X]^2_{d_0} + \sum \Real[X]^2_{d_1}e_1 + \cdots + \sum \Real[X]^2_{d_c}e_c\\
& = \left\{ \sum_{j=0}^{c} q_je_j\ \Big|\ q_j \in \sum \Real[X]^2,\ \deg(q_je_j) \leq d \right\}.
\end{split}
\end{equation}
Above, 
\begin{eqnarray}
& d \geq \max \{\deg e_1, \ldots, \deg e_c, 1\},\\
& d_j := \max \{w \in \mathbb{N}\ |\ 2w + \deg e_j \leq d \}.
\end{eqnarray}

Replacing $M$ by $M_d$, we obtain the following pair of primal-dual optimization problems
\begin{align*}
(P_d) && \text{minimize}\hspace*{0.2in}& L(X_i) &\\
&& \text{subject to}\hspace*{0.2in} & L : \Real[X]_d \to \Real \text{ is linear}, &\\
&& \hspace*{0.2in} & L(1) = 1, &\\
&& \hspace*{0.2in} & L(M_d) \subseteq [0, \infty), &\\
(D_d) && \text{maximize}\hspace*{0.2in}& a &\\
&& \text{subject to}\hspace*{0.2in} & a \in \Real, &\\
&& \hspace*{0.2in} & X_i - a \in M_d.
\end{align*}

Denoting the solution of $(P_d)$ by $P_d^*$ and that of $(D_d)$ by $D_d^*$, the Lasserre's theorem \cite{Lasserre01} guarantees that $\{D_d^*\}$ and $\{P_d^*\}$ are increasing sequences that converge to $l''_i$ and satisfy $D_d^* \leq P_d^* \leq l''_i$. This property is an important feature of our algorithm. In fact, $l''_i$ is a mere lower bound for the bounding box. The tighter the better, but it should not be overestimated as some parts of the roadmap will remain uncontained in that case. The fact that our consecutive approximations converge from below to $l''_i$ assures that $l''_i$ will never be overestimated. The analogous property guarantees that $u''_i$ will not be underestimated.

We solve these optimization problems by transforming them into semidefinite programs. We denote the set of possible monomial exponent vectors with total degree not more than $d$ by 
\begin{equation}
\mathcal{E}(d) := \{ \alpha \in \left(\mathbb{N}\cup\{0\}\right)^n\ |\ |\alpha|_1 \leq d\},
\end{equation}
the exponent vector of $X_i$ by 
\begin{equation}
\iota = (0, \ldots, \iota_i = 1, \ldots, 0),
\end{equation}  
the set of symmetric \emph{positive semidefinite} $r \times r$ matrices by
$\Real_{s+}^{r \times r}$, and the inner product of two $r \times r$ matrices $A$ and $B$ by
\begin{equation}
\langle A, B \rangle := \sum_{j,l=1}^r A(j,l)B(j,l).
\end{equation}

Define matrices $A_{\alpha j} \in \Real_{s+}^{|\mathcal{E}(d_j)| \times |\mathcal{E}(d_j)|}$ for $j \in \{0, \ldots, c\}$ and $\alpha \in \mathcal{E}(d)$ implicitly by 
\begin{equation}
X^{\beta+\gamma}e_j = \sum_{\alpha\in\mathcal{E}(d)} A_{\alpha j}(\beta, \gamma)X^\alpha,
\end{equation} 
for $\beta,\gamma \in \mathcal{E}(d_j)$. Simply, $A_{\alpha j}(\beta, \gamma)$ is the coefficient of $X^\alpha$ in $X^{\beta+\gamma}e_j$. In that case, $(P_d)$ and $(D_d)$ become the following pair of primal-dual semidefinite programs:
\begin{align*}
& (P^{\text{sdp}}_d) & \text{minimize}\hspace*{0.2in} & \sum_{j=0}^c \langle A_{0j}, Q_j \rangle &\\
&& \text{subject to}\hspace*{0.2in} & Q_j \in \Real_{s+}^{|\mathcal{E}(d_j)| \times |\mathcal{E}(d_j)|}, &\\
&&& \sum_{j=0}^c \langle A_{\iota j}, Q_j \rangle = 1, & \\
&&& \sum_{j=0}^c \langle A_{\alpha j}, Q_j \rangle = 0,\ \alpha \in \mathcal{E}(d)\backslash\{0, \iota\}, &
&\\
&(D^{\text{sdp}}_d) & \text{maximize}\hspace*{0.2in}& y_{\iota} &\\
&& \text{subject to}\hspace*{0.2in} & y_\alpha \in \Real,\ 0 \not= \alpha \in \mathcal{E}(d), &\\
&&& A_{0j} - \sum_{\alpha \in \mathcal{E}(d)\backslash\{0\}} y_\alpha A_{\alpha j} \ \text{ is positive}&\\
&&&\ \ \ \ \ \ \ \ \ \ \text{semidefinite},\ j \in \{0, \ldots, c\}.
\end{align*}

\subsubsection{Box Splitting} The algorithm splits a box through dividing its largest interval at the point that yielded the optimal value in the box shrinking optimization above if that point is in the interior. Otherwise, the algorithm splits a box through halving its largest interval.

\begin{figure*}[ht!]
\begin{center}
\begin{tabular}{cc}
\includegraphics[width=0.40\textwidth]{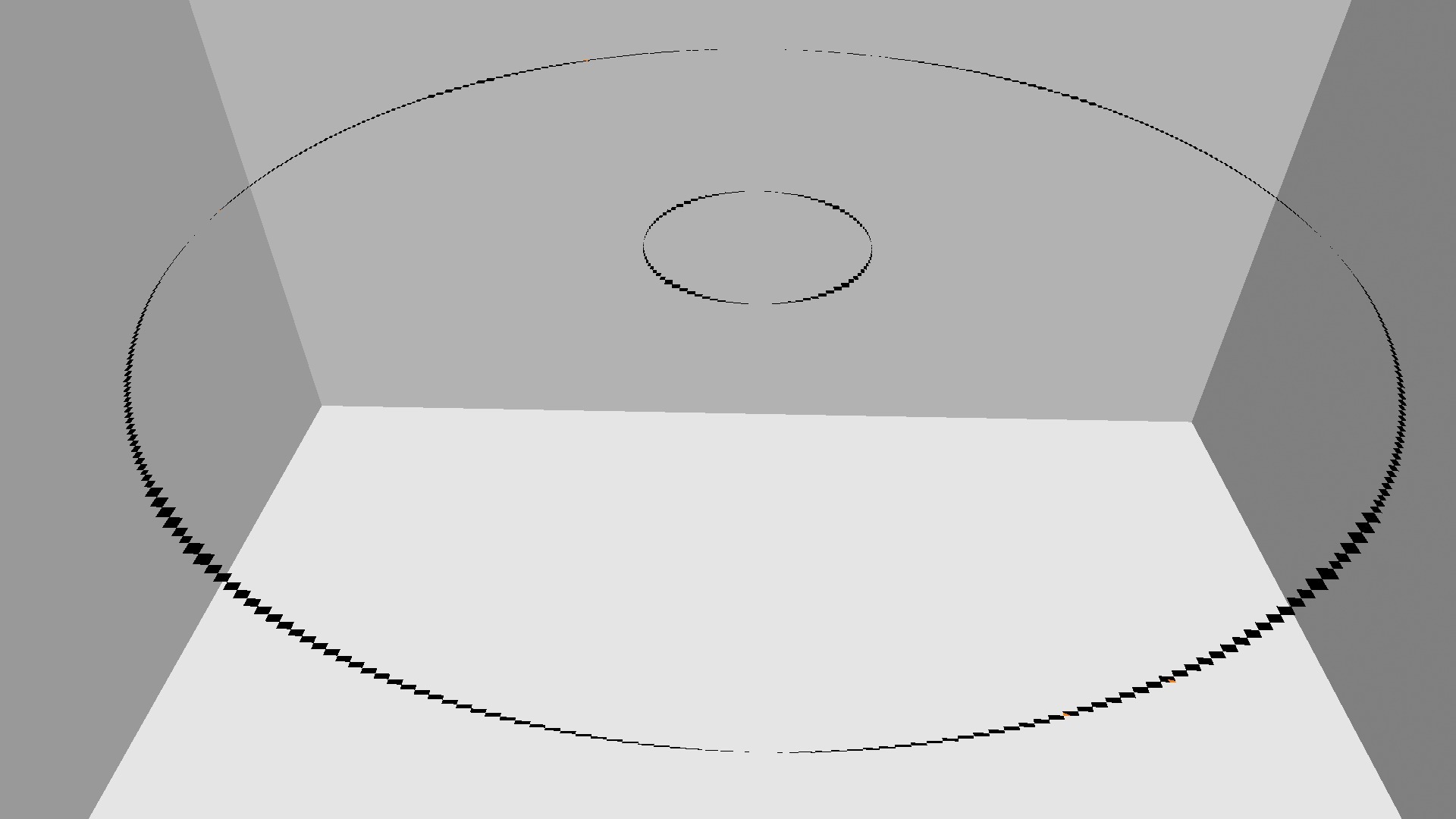} &
\includegraphics[width=0.40\textwidth]{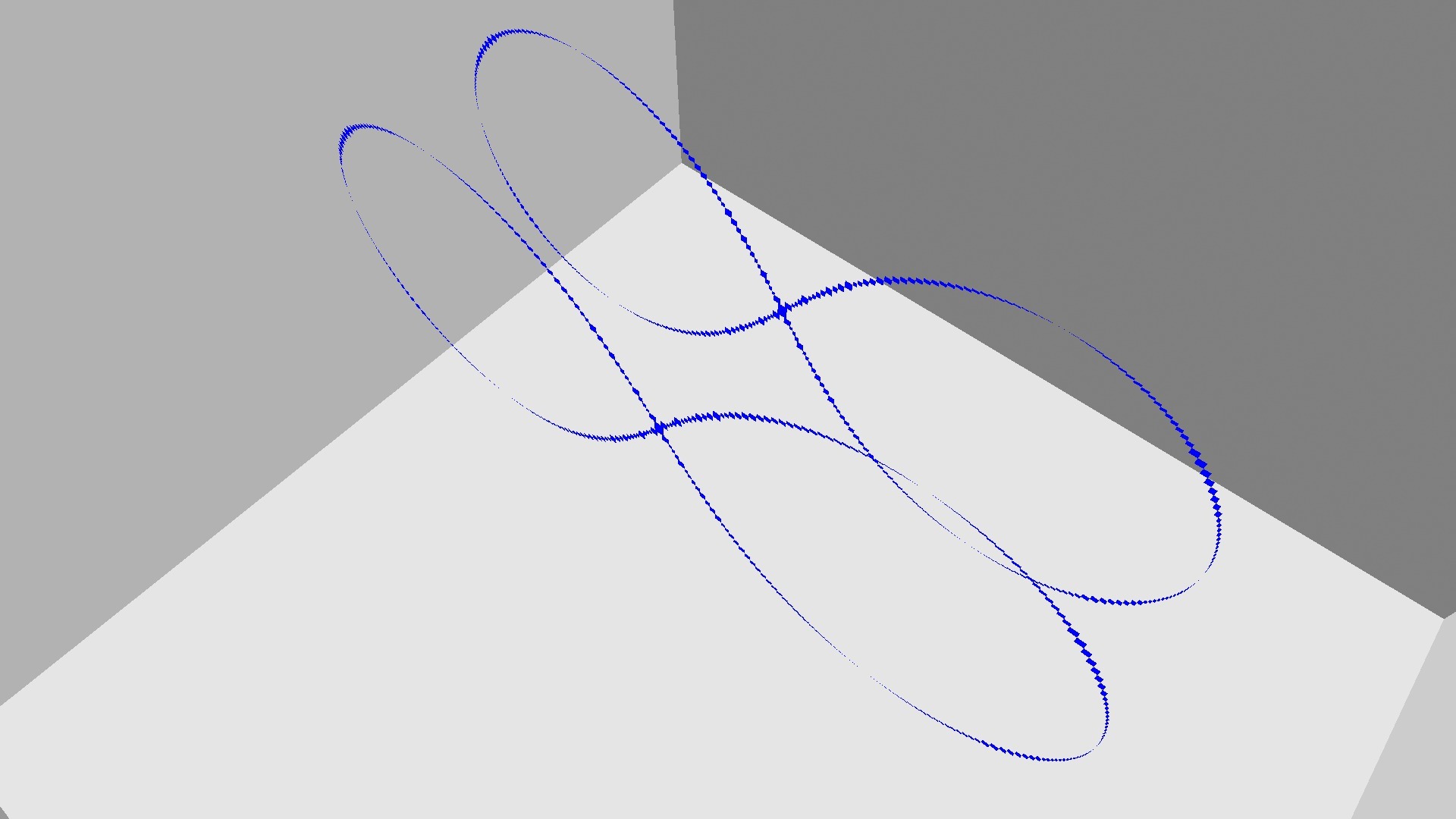} \\
Skeleton & Recursion \\
& \\
\multicolumn{2}{c}{\includegraphics[width=0.82\textwidth]{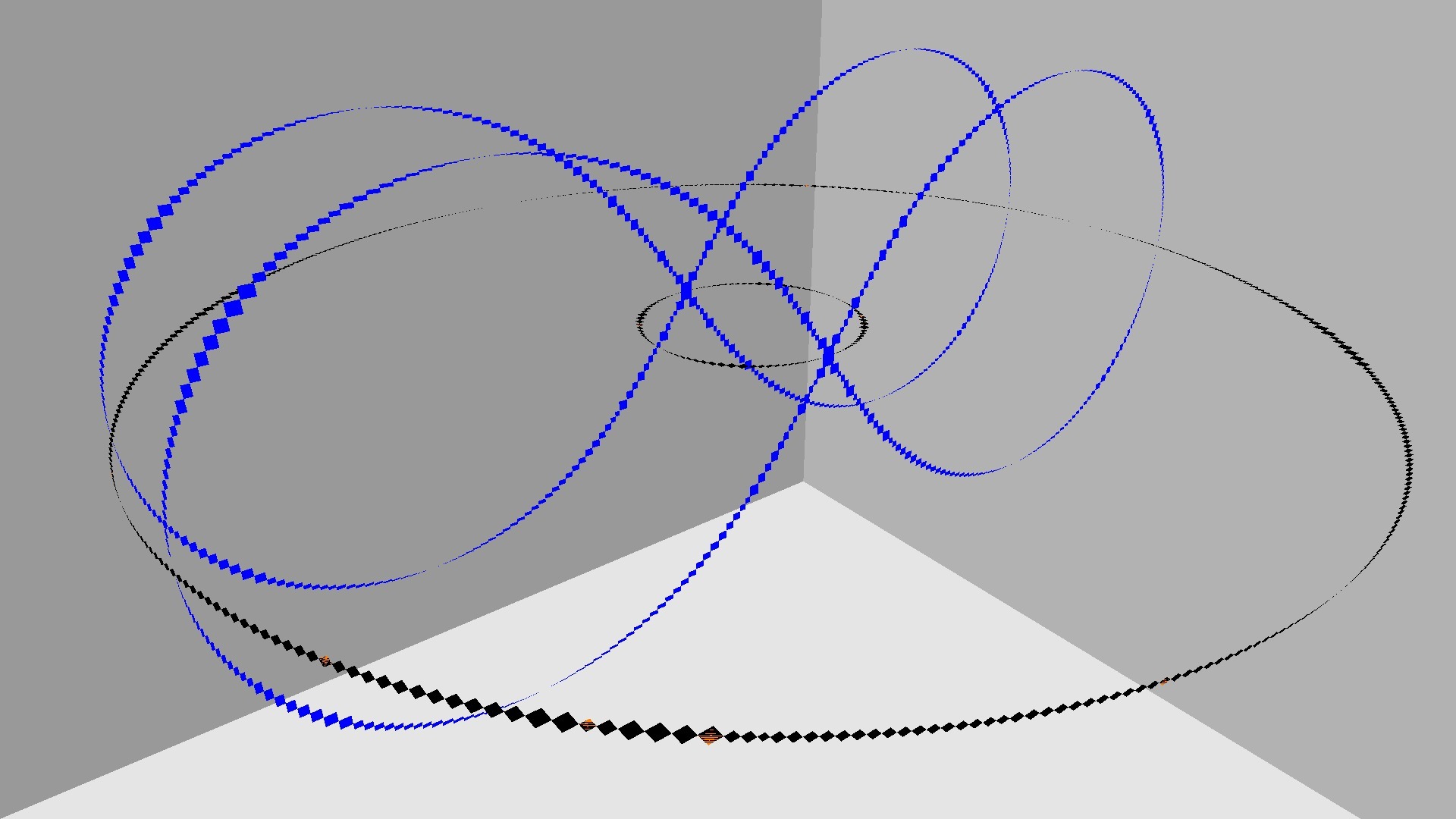}}\\
\multicolumn{2}{c}{Roadmap}\\
\end{tabular}
\end{center}
\caption{The skeleton and recursion portions of the torus roadmap (top) and the torus roadmap (bottom).}\label{fig:torus}
\end{figure*}

\begin{figure*}[ht!]
\begin{center}
\begin{tabular}{ccc}
\includegraphics[width=0.31\textwidth]{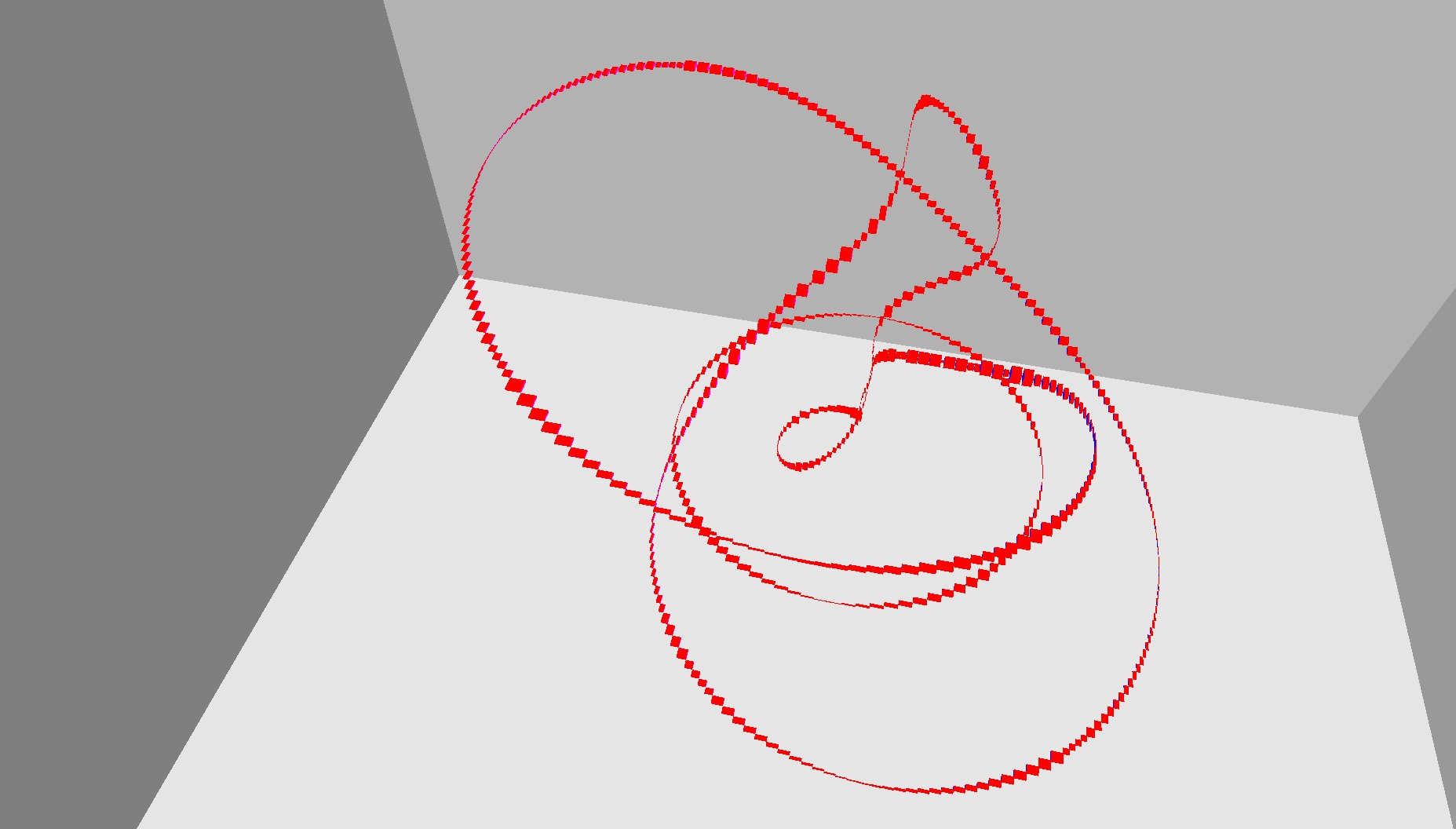} &
\includegraphics[width=0.31\textwidth]{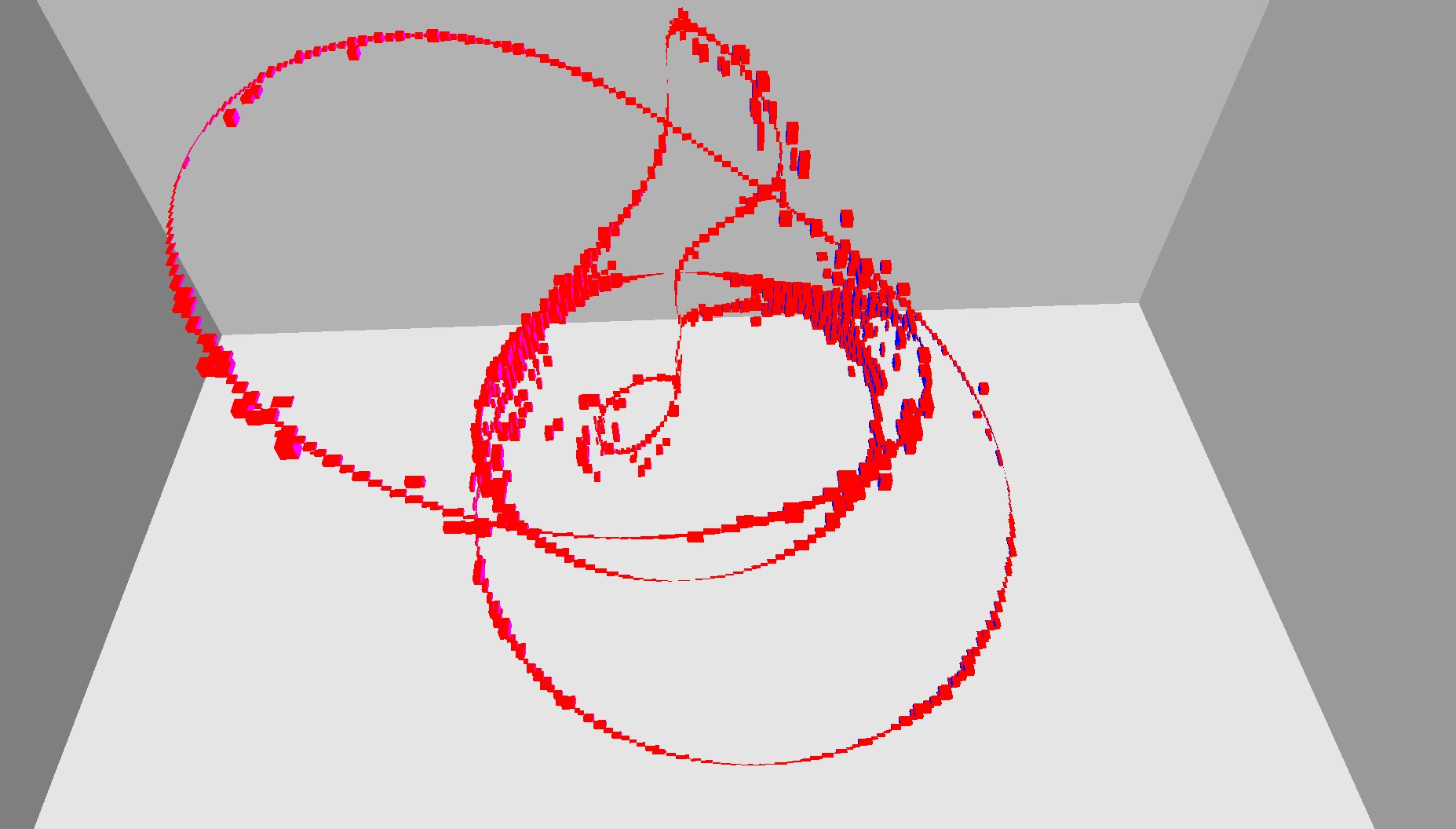} &
\includegraphics[width=0.31\textwidth]{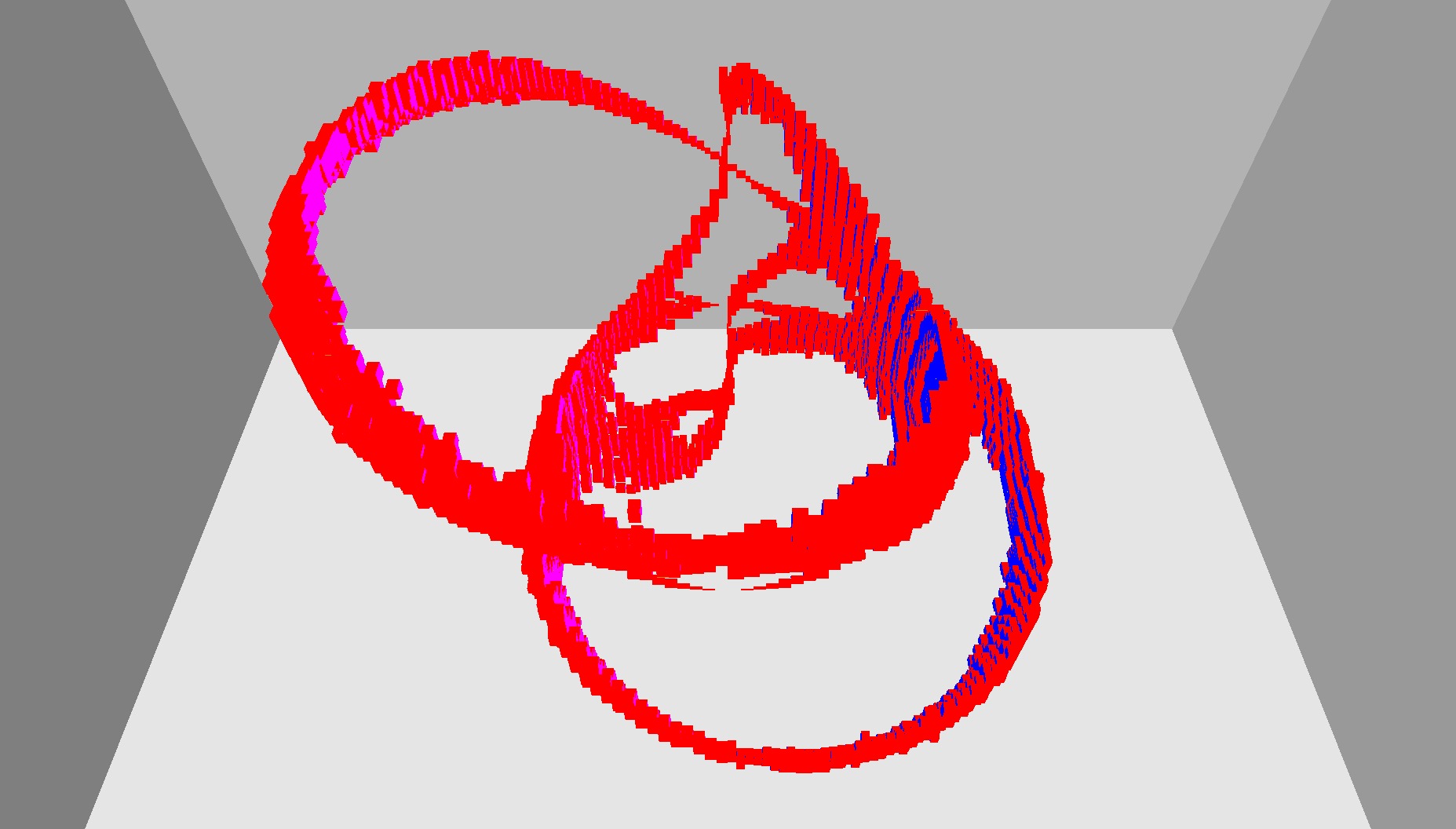} \\
NUROA & IBEX & Realpaver \\
\end{tabular}
\end{center}
\caption{\hamid{Roadmap skeleton for the Klein bottle, computed by NUROA, IBEX \cite{Araya14}, and Realpaver \cite{Granvilliers06}.}}\label{fig:klein}
\end{figure*}

\section{Results}\label{sec:results}

\subsection{Implementation}
The algorithm was written in C++. OpenGL was utilized to visualize the output of the program. For solving semidefinite programs to shrink the boxes, there are efficient open source semidefinite programming packages such as SDPA \cite{Kim11}, SDPARA \cite{Yamashita12}, SDPARA-C \cite{Nakata06}, CSDP \cite{BorYou07}, and DSDP \cite{BenYeZha00}. For convenience, we used CSDP \cite{BorYou07} in our implementation that is capable of solving problems of the form
\begin{align*}
&& \text{maximize}\hspace*{0.2in}& tr(CX) &\\
&& \text{subject to}\hspace*{0.2in} & tr(A_iX)=a_i,\ 1 \leq i \leq m &\\
&& \hspace*{0.2in} & X \in \Real_{s+}^{r \times r},
\end{align*}
where all the $A_i$, $X$, and $C$ are real and symmetric $r \times r$ matrices.

CSDP has a special, fast treatment of block diagonal matrices. Fortunately, the finalized semidefinite program in Section \ref{sec:shrinking} is in fact a program with matrices in block diagonal structure. Therefore, the primal semidefinite program, which is fed into CSDP, is
\begin{align*}
&& \text{maximize}\hspace*{0.2in} & \langle A_{0}, Q \rangle &\\
&& \text{subject to}\hspace*{0.2in} & \langle A_{\iota}, Q \rangle = 1, & \\
&& \hspace*{0.2in} & \langle A_{\alpha}, Q \rangle = 0,\ \alpha \in \mathcal{E}(d)\backslash\{0, \iota\}, & \\
&& \hspace*{0.2in} & Q \in \Real_{s+}^{r \times r},
\end{align*}
where for each $\beta \in \mathcal{E}(d)$, $A_{\beta}$ consists of $c+1$ diagonal blocks $A_{\beta j}$, $j \in \{0, \ldots, c\}$.

It has been shown that $P^*_d$ and $D^*_d$ converge rapidly to the solution in practice. Moreover, we only require a lower bound for $l''_i$ (upper bound for $u''_i$), not necessarily the optimal value. Hence, although the size of these programs are exponential in the dimension, i.e. $|\mathcal{E}(d)|$ is $O(d^n)$, we expect to solve these optimizations only for few small $d$. Moreover, as mentioned before, $A_{\beta j}$ are sparse in practice and may be grouped to simplify these programs using the special structure of (\ref{equ:g}), (\ref{equ:skeleton}), and (\ref{equ:e0})-(\ref{equ:e6}).

\subsection{Experiments}

\begin{figure}
\begin{center}
\includegraphics[width=0.45\textwidth]{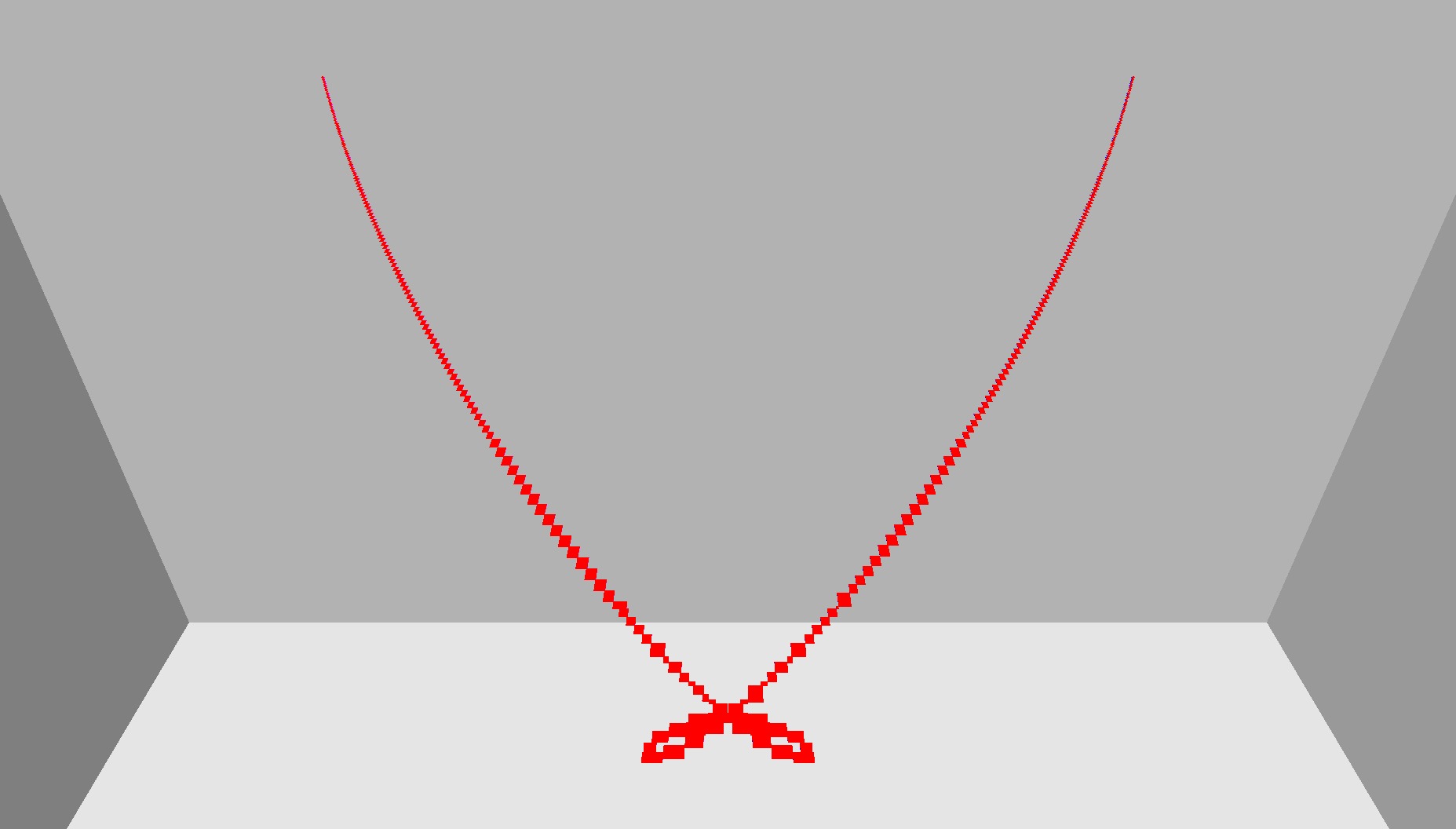} \\
Bow curve \\\ \\
\includegraphics[width=0.45\textwidth]{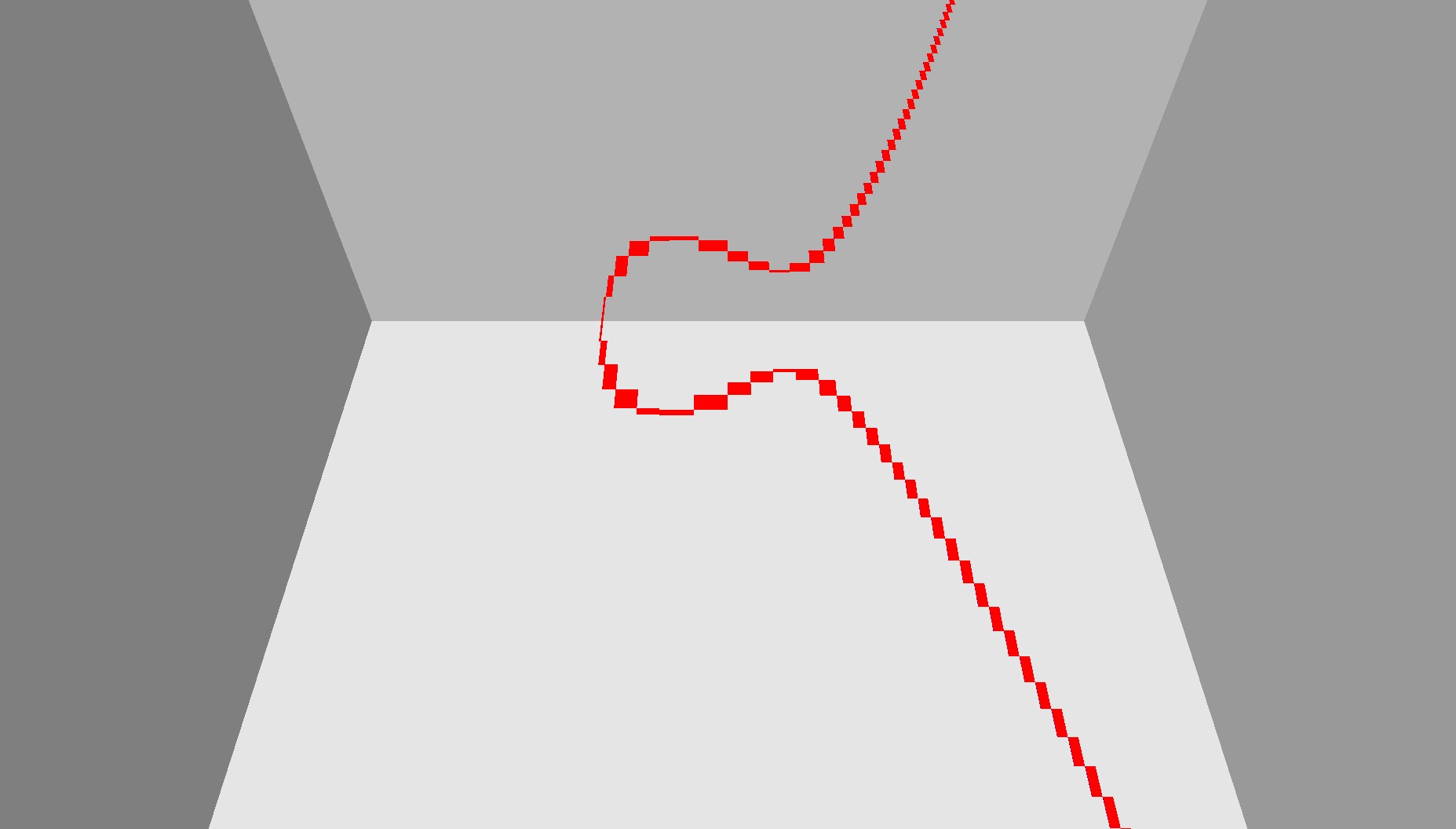} \\
Elliptic curve \\\ \\
\includegraphics[width=0.45\textwidth]{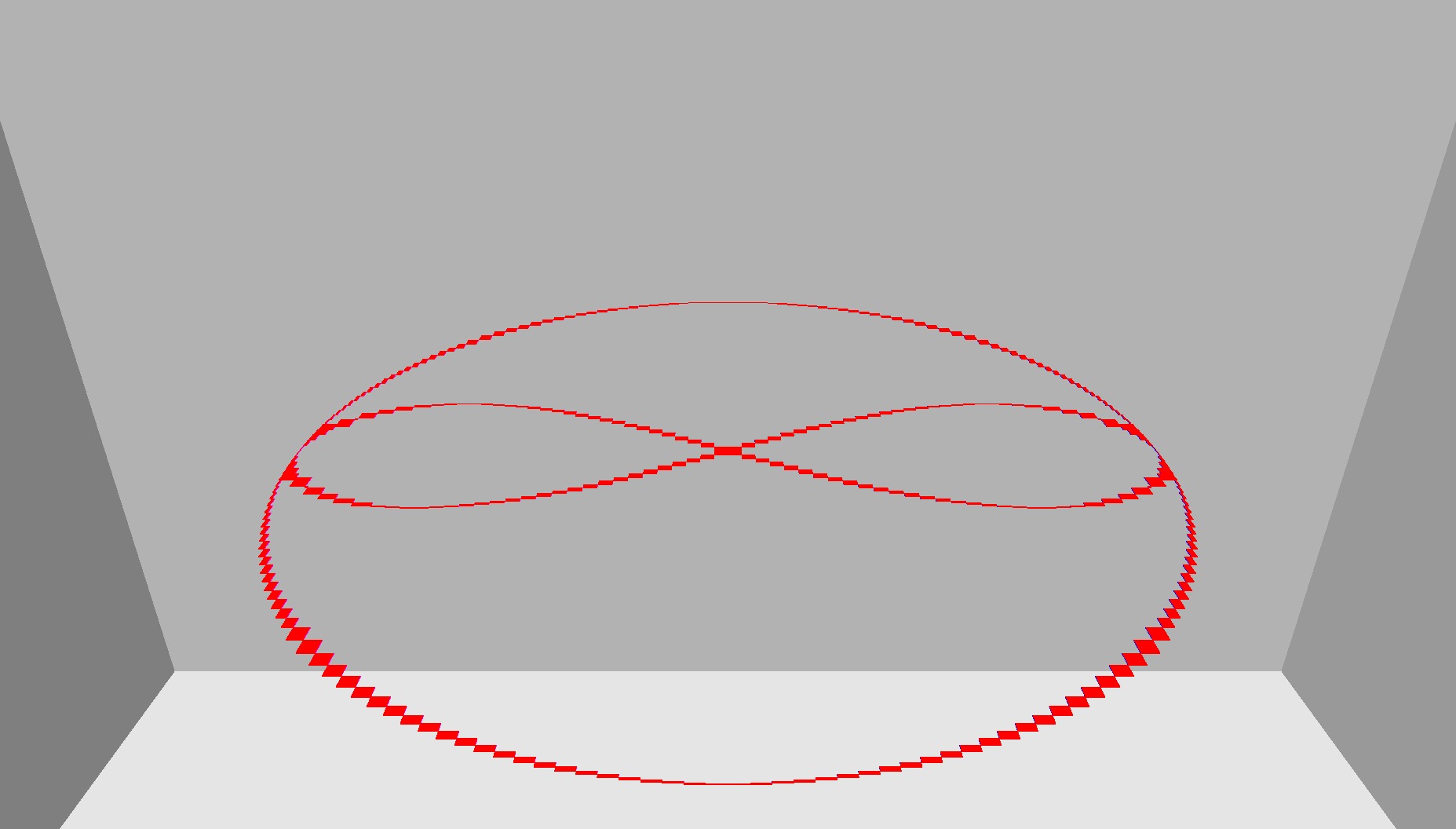} \\
Watt's curve
\end{center}
\caption{\hamid{The bow, elliptic, and Watt's curves computed by NUROA.}}\label{fig:bow-ell-watts}
\end{figure}

\hamid{We designed four test cases: (i) three-leaf clover planar curve embedded in $\Real^3$, (ii) the Canny's roadmap for the torus, (iii) the roadmap skeleton for the Klein bottle, and (iv) the sphere skeleton, bow, elliptic, and Watt's curves embedded in $\Real^3$.} Fig. \ref{fig:clover} depicts the boxes that enclose the clover curve, and Fig. \ref{fig:torus} shows the boxes that contain the Canny's roadmap skeleton, recursion, and their union. In both test cases, boxes were refined until their longest side was no longer than $0.1$. \hamid{Fig. \ref{fig:klein} shows the Klein bottle roadmap skeleton computed by three tools: (i) our tool NUROA, (ii) IBEX \cite{Araya14}, and (iii) Realpaver \cite{Granvilliers06}. As it can be observed from the figure and the number of boxes in Table \ref{tab:results1}, IBEX and Realpaver results were not accurate enough for path planning. Fig. \ref{fig:bow-ell-watts} illustrates NUROA results for the bow, elliptic, and Watt's curves.}

The three-leaf clover curve, defined in (\ref{equ:clover}), was used to confirm that our algorithm works correctly on any algebraic curve, the special case of which is Canny's roadmap skeleton. The algorithm was able to capture its connectivity.
\begin{equation}\label{equ:clover}
(x^2 + y^2)^2 - x^3 + 3xy^2 = 0.
\end{equation}

Canny's standard example in his dissertation was the torus roadmap \cite{Can88}. We chose that as the second test case. In this case too, the roadmap connectivity was preserved. The considered torus was
\begin{equation}\label{equ:torus}
36(x^2 + y^2) - (5+x^2+y^2+z^2)^2 = 0,
\end{equation}
with radii $3$ and $2$.

\hamid{Since the Klein bottle, 
\begin{equation}
\begin{aligned}
& (x^2+y^2+z^2+2y-1) \left[(x^2+y^2+z^2-2y-1)^2-8z^2\right] \\
& + 16xz(x^2+y^2+z^2-2y-1) = 0,
\end{aligned}
\end{equation}
has a relatively complex silhouette, we chose it for comparison between our tool and competitors, IBEX and Realpaver. NUROA clearly outperforms competitors in terms of the number of boxes and accuracy (see Table \ref{tab:results1} and Fig. \ref{fig:klein}).}

We ran our code on a single AMD Opteron 6180 SE 2.5 GHz core. Tables \ref{tab:results1} and \ref{tab:results2} show the number of boxes and running time in each case.
\begin{table}[h!]
\caption{\hamid{The number of boxes for the clover curve, torus skeleton and recursion, the Klein bottle skeleton, sphere, bow, elliptic, and Watt's curves computed by NUROA, IBEX, and Realpaver with precisions $0.5$ and $0.1$. Below, $d$ is the degree of monomials in the finite module approximation $M_d$ in the Lasserre's approach.}}\label{tab:results1}
\begin{center}
\begin{tabular}{|c|c|c|c|c|c|c|c|}
\hline
\multirow{2}{*}{Case} & \multicolumn{3}{c|}{NUROA} & \multicolumn{2}{c|}{IBEX} & \multicolumn{2}{c|}{Realpaver}\\
\cline{2-8}
 & $d$ & 0.5 & 0.1 & 0.5 & 0.1 & 0.5 & 0.1 \\
\hline
\hline
Clover & 5 & {\bf 10} & {\bf 94} & 26 & 204 & 18 & 152\\
Torus & 5 & {\bf 108} & {\bf 492} & 224 & 1028 & 152 & 768 \\
Klein & 8 & {\bf 120} & {\bf 524} & 1712 & 5478 & 1000 & 6016\\
Sphere & 5 & 16 & {\bf 96} & 24 & 162 & {\bf 12} & {\bf 96}\\
Bow & 8 & {\bf 50} & {\bf 186} & 242 & 1194 & 108 & 710\\
Elliptic & 8 & {\bf 64} & {\bf 276} & 268 & 1338 & 146 & 618\\
Watt's & 8 & {\bf 56} & {\bf 312} & 108 & 544 & 60 & 340 \\
\hline
\end{tabular}
\end{center}
\end{table}

\begin{table}[h!]
\caption{\hamid{The running time (s) for the clover curve, torus skeleton and recursion, the Klein bottle skeleton, sphere, bow, elliptic, and Watt's curves computed by NUROA, IBEX, and Realpaver with precisions $0.5$ and $0.1$. Below, $d$ is the degree of monomials in the finite module approximation $M_d$ in the Lasserre's approach.}}\label{tab:results2}
\begin{center}
\begin{tabular}{|c|c|c|c|c|c|c|c|}
\hline
\multirow{2}{*}{Case} & \multicolumn{3}{c|}{NUROA} & \multicolumn{2}{c|}{IBEX} & \multicolumn{2}{c|}{Realpaver}\\
\cline{2-8}
 & $d$ & 0.5 & 0.1 & 0.5 & 0.1 & 0.5 & 0.1 \\
\hline
\hline
Clover & 5 & 6.13 & 66.85 & 0.02 & 0.1 & 0.01 & 0.05\\
Torus & 5 & 56.76 & 239.49 & 0.04 & 0.23 & 0.08 & 0.27\\
Klein & 8 & 1082.38 & 4494.94 & 1.18 & 4.89 & 0.77 & 5.89\\
Sphere & 5 & 7.82 & 45.82 & 0.004 & 0.02 & 0.00 & 0.00\\
Bow & 8 & 596.2 & 1753.73 & 0.06 & 0.28 & 0.02 & 0.14\\
Elliptic & 8 & 599.34 & 2130.94 & 0.07 & 0.26 & 0.01 & 0.06\\
Watt's & 8 & 332.85 & 1820.55 & 0.04 & 0.14 & 0.01 & 0.06\\
\hline
\end{tabular}
\end{center}
\end{table}
\hamid{It is clear from Tables \ref{tab:results1} and \ref{tab:results2} that IBEX and Realpaver compromise accuracy to achieve significant speedup, in comparison to NUROA.} However, the running time of NUROA reported in Table \ref{tab:results2} is without any effort to parallelize the code. Using GPUs, we expect to achieve a significant speedup  as the computations for each box are independent from those for another box, and hence, NUROA can be parallelized in a relatively straight forward manner. 

\section{Discussion and Conclusions}\label{sec:conclusion}
\hamid{We demonstrated that ease of implementation can be brought to the theoretically well-rooted, elegant world of complete, combinatorial motion planning approaches that have suffered so far from implementation and running time sophistications. Our proposed methodology, NUROA, was compared with IBEX and Realpaver in a number of experiments. Simulation results suggest that, unlike IBEX and Realpaver, NUROA does not compromise accuracy in favor of speed. Moreover,}
\begin{itemize}
  \item NUROA can be customized for a particular accuracy and computational intensity requirements by setting the resolution parameter: the higher the resolution, the more details of roadmap is captured.
  \item NUROA can be highly parallelized, particularly on GPUs, which makes it deployable on embedded devices.  
\end{itemize}

\hamid{Actual practical applicability of the proposed planner can be inferred from the results of the simple examples that we presented in this paper. Essentially, the torus roadmap was captured with a few hundred boxes. Therefore, the entire computation can be done on a GPU in the GPU memory without the need for data transfer to/from the main memory. On a GPU, we expect the torus roadmap to be computed in a fraction of a second.}

\hamid{The main weakness of probabilistic approaches is dealing with narrow passages. Our proposed approach is favorable in that case in comparison with probabilistic approaches because the purely symbolic roadmap algorithm is not sensitive to geometry at all. Albeit unlike the purely symbolic case, NUROA deals with the geometry to some extent through controlling the sizes of the enclosing boxes. However, NUROA does not miss a narrow passage but it may incorrectly capture a narrow obstacle as free space. Overall, NUROA's convergence rate is expected to be higher than that of probabilistic approaches.}

\bibliography{masterref,pub,MyPubs}
\bibliographystyle{plain}

\end{document}